\documentclass[runningheads]{llncs}

\usepackage[final,year=2026]{eccv}
\usepackage{eccvabbrv}
\usepackage{lmodern}

\usepackage{graphicx}
\usepackage{booktabs}
\usepackage[accsupp]{axessibility}

\usepackage[breaklinks,colorlinks,citecolor=eccvblue]{hyperref}
\usepackage{orcidlink}

\usepackage[capitalize]{cleveref}
\crefname{section}{Sec.}{Secs.}
\Crefname{section}{Section}{Sections}
\crefname{figure}{Fig.}{Figs.}
\Crefname{figure}{Figure}{Figures}
\crefname{table}{Tab.}{Tabs.}
\Crefname{table}{Table}{Tables}
\crefname{equation}{Eq.}{Eqs.}

\usepackage{amsmath,amssymb}

\usepackage{booktabs}
\usepackage{multirow}
\usepackage{array}
\usepackage{colortbl}

\usepackage{graphicx}
\usepackage{tikz}
\usepackage{pgfplots}
\pgfplotsset{compat=1.17}
\usepgfplotslibrary{groupplots}
\usetikzlibrary{patterns,positioning}
\usepackage{pgfplotstable}

\usepackage{cuted}
\usepackage{xspace}
\usepackage{enumitem}
\setlist{nosep,leftmargin=1.4em}

\newcommand{\epdms}{\textup{EPDMS}\xspace}

\newcommand{\pdmc}{\textup{PDM-Closed}\xspace}
\newcommand{\ltfv}{\textup{LTFV6}\xspace}
\newcommand{\ddrive}{\textup{DiffusionDrive}\xspace}
\newcommand{\ignoreall}{\textup{Ignore-All}\xspace}
\newcommand{\humanreplay}{human replay\xspace}

\newcommand{\IA}{\textup{IA}}
\definecolor{hilite}{rgb}{0.88,0.93,0.98}
\definecolor{winred}{rgb}{0.80,0.15,0.15}
\definecolor{wingreen}{rgb}{0.10,0.45,0.20}
\definecolor{figorange}{rgb}{0.847,0.690,0.498}
\definecolor{cvprblue}{rgb}{0.21,0.49,0.74}
\definecolor{figgreen}{rgb}{0.518,0.612,0.553}

\title{When Shared Rollouts Fail in Defensive Driving Evaluation:\texorpdfstring{\\}{ }
A NAVSIM Score Basis Audit}
\titlerunning{NAVSIM Score Basis Audit}

\author{Ziang Wei\orcidlink{0000-0002-5015-8696} \and Minjun Yu\orcidlink{0009-0002-6589-0342} \and Zheyuan Lai\orcidlink{0009-0008-4694-8573}  \and Mingjie Pang\orcidlink{0000-0002-0321-646X}  \and Wei Li\orcidlink{0000-0002-0059-3745}}
\authorrunning{Z. Wei and M. Yu and Z. Lai and M. Pang and W. Li}
\institute{EABOT.AI}

\hypersetup{
  pdftitle={When Shared Rollouts Fail in Defensive Driving Evaluation: A NAVSIM Score Basis Audit},
  pdfauthor={Ziang Wei, Minjun Yu, Zheyuan Lai, Mingjie Pang, Wei Li},
  pdfsubject={Defensive driving benchmark auditing},
  pdfkeywords={defensive driving evaluation, benchmark auditing, reference-conditioned scoring, actor-blind probes, numerical robustness}
}

\begin{document}
\maketitle

\begin{abstract}
Defensive driving scores are useful only when they preserve distinctions between
policies that observe surrounding actors and those that do not. Re-simulation
benchmarks may use reference-conditioned forgiveness, under which an agent
receives credit when the logged human reference fails a compliance channel. When
agent and reference share an unstable rollout transformation, this rule can
propagate shared reference failures into broad compliance credit.

We audit this risk in NAVSIM v2.2 original scene single-stage scoring. Under
the affected documented-stack condition on the audited numerical backend, the
route-blind \ignoreall probe and a route-aware actor-blind probe outrank human
replay and \pdmc over the complete 12,146-token navtest split. A fresh
installation following the public specification reproduces rollout divergence
on a fixed 32-token diagnostic set. A same-source dependency stack control and
an exact-input diagnostic isolate dependency-sensitive numerical behavior in the
shared velocity refit. On a 450-token control pool, replacing only the solver
eliminates rollout divergence and restores blind-last ordering while keeping
forgiveness enabled. Thus, the numerical instability is the direct trigger.
Reference-conditioned forgiveness propagates the resulting shared reference
failures into compliance credit. We contribute an audit protocol requiring
score basis and stack disclosure, blind probes, overwrite reporting, and
rollout stability tests before using such scores for defensive driving claims.
\end{abstract}

\begin{figure*}[t]
\centering
\includegraphics[width=0.98\linewidth]{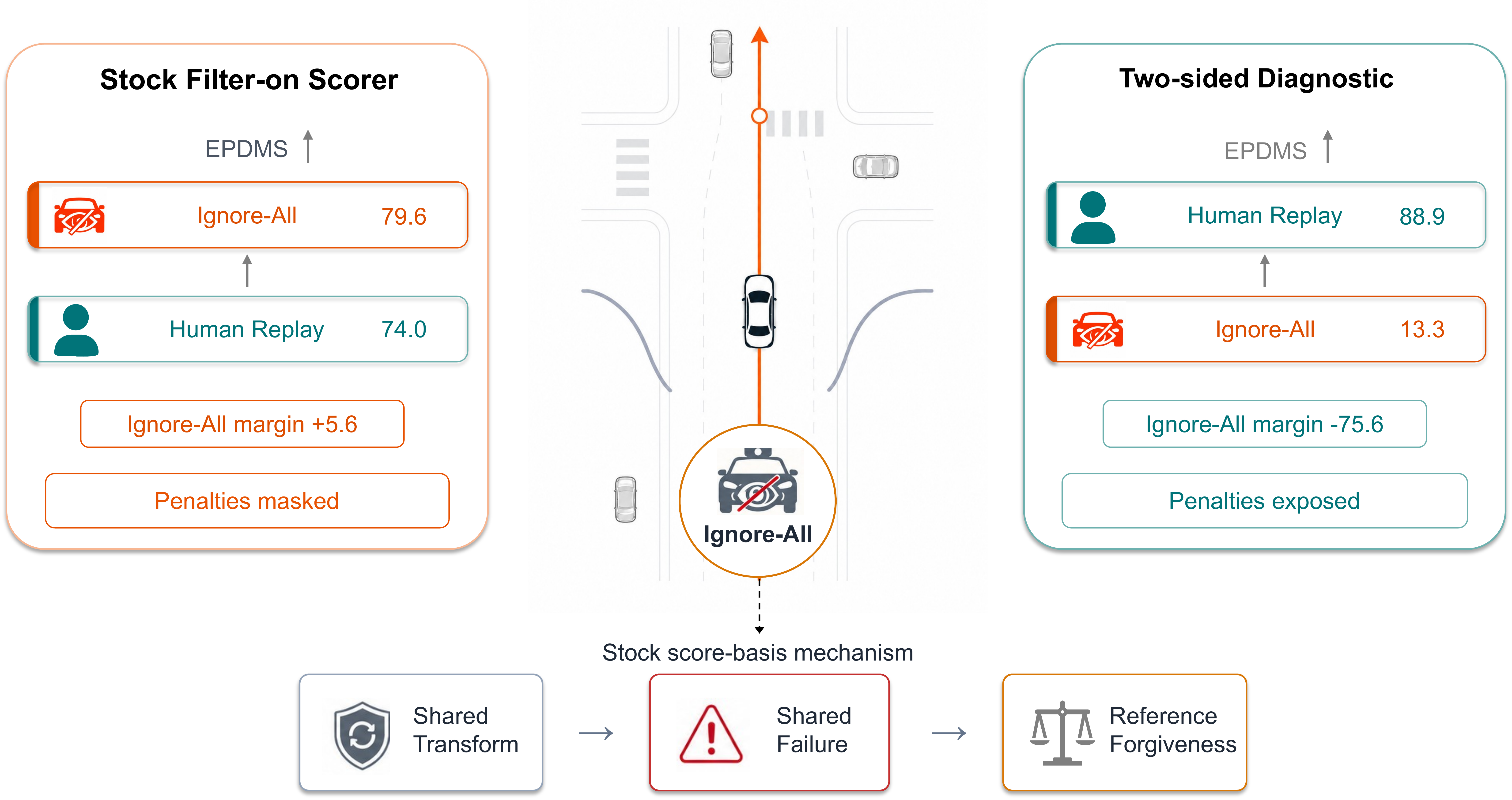}
\caption{\textbf{Shared-failure forgiveness collapse.}
Under stock filter-on scoring in the affected documented-stack condition
(left), the actor-blind \ignoreall, which receives no surrounding-actor
input, outranks human replay with 79.6 vs.\ 74.0 Extended Predictive Driver
Model Score (EPDMS). Shared reference failure produces near-saturated
fail-to-pass overwrite in the core Drivable Area Compliance (DAC), Driving
Direction Compliance (DDC), and lane keeping (LK) channels before aggregation.
A two-sided finite-difference
diagnostic that bypasses the unstable shared refit on both the proposal and
reference sides reverses the pairwise ordering (human replay 88.9 vs.\
\ignoreall 13.3 EPDMS). Because proposal and reference share the transform,
common numerical failures trigger policy-level overwrite. Full results appear in
\Cref{tab:full-navtest,tab:score-basis-disclosure,tab:twosided}.}
\label{fig:teaser}
\end{figure*}

\section{Introduction}
\label{sec:intro}

Defensive driving evaluation is commonly interpreted as evidence that a planner
responds to surrounding actors and interaction risks rather than merely matching
a nominal trajectory. That inference requires a basic ranking property:
an actor-blind diagnostic should not systematically outrank strong
actor-aware comparators under a score interpreted as evidence of
actor-sensitive driving. Large-scale
driving benchmarks increasingly use \emph{re-simulation}: a planner proposes an
ego trajectory, the scene is rolled forward, and safety and quality terms are
scored against a logged human reference~\cite{dauner2023pdm,dauner2024navsim,cao2025pseudosimulation}.
Such scores support claims about perception, 3D representation, and multimodal
reasoning~\cite{hu2023uniad,jiang2023vad,hwang2024emma,sima2024drivelm}. Those
claims are credible only if the score preserves a necessary distinction between
actor-aware comparators and actor-blind probes.

We audit a structural validity risk in this documented NAVSIM scorer configuration.
It uses reference-conditioned forgiveness. If the logged human reference fails a
compliance channel, the agent may be forgiven on it. The rule avoids penalizing
agents for violations also attributed to the logged reference.
The dependency-sensitive numerical
instability in the shared rollout and refit path is the proximate trigger in our
audit. Because proposal and reference traverse that same path, forgiveness can
propagate the resulting shared reference failures into credit and erase
behavior-relevant distinctions. We call this \emph{shared-failure forgiveness
collapse}. It requires a transformation shared by proposal and reference,
reference-conditioned forgiveness, and broad shared failure on channels relevant
to safe driving. We use \emph{score basis} for the complete configuration,
including rollout transformation and forgiveness.

Blind policies provide a lightweight diagnostic. \ignoreall is route-blind and
fixed-straight. The route-aware actor-blind probe follows declared mission
roadblocks using current ego state and static-map lane and lane-connector
baselines only. It uses no actor boxes, tracks, trajectories, collision
predictions, or logged future ego trajectory. If either probe outranks
actor-aware comparators, the score fails this actor-blind sanity check. This is
a falsification test of a necessary condition, not a complete measure of
defensive driving competence.

We instantiate the audit on NAVSIM v2.2 original scene single-stage scoring.
Under the affected documented-stack condition, both probes outrank
\humanreplay and \pdmc on the frozen 12,146-token navtest split. Yet a two-sided diagnostic places human replay and \pdmc above the route-aware actor-blind probe, which in turn ranks above the route-blind probe,
separating route following from actor blindness. The direct trigger is
dependency-sensitive instability in the stock velocity refit. The default
pseudoinverse can produce a kilometer-scale off-road rollout. The logged
reference traverses the same refit, activating forgiveness on almost every token.

Two-sided, fallback, solver, and dependency-stack controls localize the unstable
shared rollout basis. With forgiveness enabled, solver replacement eliminates
divergence and restores blind-last ordering on the 450-token control pool. A
fresh installation reproduces scorer divergence on a fixed 32-token diagnostic
set, while an exact-input diagnostic separates the affected and control stacks
without changing scorer source (\cref{sec:audit}). These are diagnostic controls,
not a replacement metric. Our contributions are as follows:
\begin{enumerate}
    \item We define shared-failure forgiveness collapse and a blind probe audit
    for defensive driving scores used to support actor-sensitive claims.
    \item We show that the affected documented-stack condition ranks both route-blind and
    route-aware actor-blind probes above actor-aware comparators, and localize a
    dependency-sensitive velocity-refit trigger.
    \item We validate the mechanism with complementary controls and derive
    dependency disclosure, overwrite reporting, blind probe, and rollout stability
    requirements for reference-conditioned re-simulation metrics.
\end{enumerate}

\section{Related Work}
\label{sec:related}

\subsection{Driving datasets, simulators, and benchmark protocols}
Autonomous driving evaluation has progressed from offline perception and
forecasting datasets toward planners tested through closed-loop or
pseudo-simulation protocols. NAVSIM and PDM-style metrics roll out proposed ego
trajectories over real-scene abstractions and aggregate progress, comfort,
collision, time-to-collision, and map compliance terms at scale
\cite{dauner2023pdm,dauner2024navsim,cao2025pseudosimulation}. Closed-loop
benchmarks such as nuPlan instead expose the planner to evolving simulated
traffic~\cite{caesar2021nuplan}, while cross-benchmark studies ask whether
offline and pseudo-simulation scores predict closed-loop performance
\cite{li2025crossbenchmark}. We ask a narrower prerequisite question: whether a
re-simulation score preserves the actor-sensitive distinctions needed to support
defensive driving claims.

\subsection{Defensive driving evaluation and actor-sensitive behavior}
Defensive driving is not equivalent to avoiding a recorded collision. It
requires anticipating latent hazards, preserving safety margins, yielding when
another actor may enter the ego path, and maintaining route and map constraints
as interactions evolve. Interaction-rich datasets make these behaviors
measurable~\cite{zhan2019interaction}. Planning benchmarks then summarize them
through safety, progress, and compliance terms
\cite{caesar2021nuplan,dauner2024navsim}. Such aggregates are routinely read as
evidence of hazard sensitivity rather than trajectory imitation alone.

A score used to support actor-sensitive claims must at least distinguish policies
that observe surrounding actors from policies that do not. Blind-last ordering alone
does not certify anticipation, comfort, or deployment safety. However, a blind
win is an important warning sign. In our audit, it becomes diagnostic because the scorer still
observes the actors, core failures are broadly overwritten, and the ordering reverses
under independently stabilized score bases.

\subsection{End-to-end driving and perception-rich planning}
End-to-end driving systems connect perception, prediction, and planning in a
learned stack. Conditional imitation, privileged teaching, and
world-on-rails supervision established strong CARLA baselines
\cite{codevilla2018cil,chen2020lbc,chen2021worldonrails}. Later systems combine
camera--LiDAR fusion~\cite{prakash2021mft,chitta2023transfuser}, explicit
interaction-aware planning~\cite{chen2022lav,wu2022tcp,shao2023interfuser}, or
object-centric representations~\cite{renz2022plant}. BEV perception, occupancy,
vectorized planning, and language-conditioned driving further expand the
behavioral claims attached to aggregate planning scores
\cite{philion2020lift,li2024bevformer,casas2021mp3,hu2023uniad,jiang2023vad,sima2024drivelm,hwang2024emma}.
NAVSIM-oriented planners such as Hydra-MDP~\cite{li2024hydramdp} and
DiffusionDrive~\cite{liao2025diffusiondrive} use pseudo-simulation as an
evaluation target. Our concern therefore precedes model comparison and asks whether the target retains the actor-sensitive information these systems are designed to exploit.

\subsection{Evaluation shortcuts, reward misspecification, and metric audits}
Driving metrics can reward shortcuts unrelated to the intended capability.
AD-MLP and BEV-Planner show that ego-status or no-perception policies can appear
competitive under open-loop nuScenes metrics~\cite{zhai2023admlp,li2024bevplanner},
and hidden-bias analyses identify related brittleness in end-to-end driving
models~\cite{jaeger2023hidden}. Those results mainly expose dataset or open-loop
metric shortcuts. Our case differs because the blind probe has poor open-loop
trajectory match, yet pseudo-simulation ranks it highest when a shared
rollout failure activates reference-conditioned forgiveness. The failure arises within the score basis rather than from exploiting ego status
or the logged future.

More generally, Goodhart- and Campbell-style effects describe how optimized
proxies can diverge from their intended meaning
\cite{campbell1979assessing,manheim2018goodhart}, while
research on reward hacking and reward misspecification examines failures of proxy
objectives~\cite{amodei2016concrete,skalse2022reward,pan2022rewardmisspecification,hendrycks2021unsolved}.
Bisimulation and value-equivalence characterize when states or models can be
treated alike for control~\cite{ferns2004bisimulation,grimm2020value,zhang2021dbc}.
Shared-failure forgiveness collapse violates the corresponding measurement
principle by mapping behaviorally different structural failures into a common
pass state before aggregation. Unlike an agent that deliberately learns to game
a reward, the blind probe here exposes an evaluation-side equivalence introduced
by the scorer itself.

\section{Audit Protocol}
\label{sec:audit}

Our reusable audit begins with a behavioral symptom and then tests its score
basis. We ask whether a blind policy can win, whether the scorer still observes
the true scene, whether reference forgiveness overwrites behavior-relevant
failures, and whether removing the shared failure restores blind-last ordering.
Together these checks separate an opaque ranking anomaly from a score basis
mechanism.

\subsection{What the audit tests}
The four questions are deliberately ordered. First, a blind win signals failure
of the actor-blind sanity check. Second, visibility checks distinguish a score basis
failure from a simpler implementation error in which actors were never passed to
the scorer. Third, channel-level masks show whether the symptom is caused by a
reference-conditioned rewrite rather than an ordinary aggregation trade-off.
Finally, interventions on the common transformation test whether that rewrite
depends on shared rollout failure. A blind win without the latter three
signatures requires a different diagnosis.

\begin{table}[!htbp]
\centering
\small
\setlength{\tabcolsep}{3.2pt}
\caption{\textbf{Observable checks in the proposed score basis audit.}
Each step distinguishes a behavioral symptom from a scorer-side mechanism; a
blind win alone is not sufficient to diagnose forgiveness collapse.}
\label{tab:audit-checklist}
\begin{tabular}{@{}>{\raggedright\arraybackslash}p{0.31\textwidth}
                    >{\raggedright\arraybackslash}p{0.33\textwidth}
                    >{\raggedright\arraybackslash}p{0.28\textwidth}@{}}
\toprule
Audit question & Observable & Failure signal \\
\midrule
Blind policy can win? & Paired blind--comparator margin & Blind $>$ actor-aware comparator \\
Scorer observes actors? & Collision-risk score response & Change with true actors \\
Reference failure rewrites? & Failure masks; fail-to-pass counts & Broad core overwrite \\
Shared rollout stable? & Fitted state; step; path deviation & Impossible state or divergence \\
Independent basis restores order? & Symmetric, solver, and stack controls & Consistent blind demotion \\
\bottomrule
\end{tabular}
\end{table}

\subsection{Scoring bases and controls}
The \emph{affected documented-stack condition} follows NAVSIM v2.2 public
commit \texttt{0a380a9} on our audited x86-64 backend with OpenBLAS (Python 3.9.23,
NumPy 1.23.4, SciPy 1.13.1, and OpenCV 4.9.0). This label identifies the
measured condition, not reproduction across hardware or linear algebra backends.
Its default
original scene scorer uses the stock pseudoinverse refit and reference-conditioned
forgiveness (\emph{stock filter-on}); \emph{stock filter-off} disables only the
forgiveness branch.

We separate the stack from the scorer path because a solver-only result cannot
by itself distinguish numerical behavior from scorer design. The \emph{same-source
control stack} changes binary dependencies only as needed to import and run the
identical scorer source (NumPy 2.0.2 and OpenCV 4.13.0; OpenBLAS 0.3.27 versus
0.3.20 in the affected condition). This serves as an operational control rather
than a production recommendation; source, dependencies, and backend are specified here.

Our primary audit targets original scene single-stage scoring. The 450-token
 solver pool is the complete original scene stage-1 set of the official
navhard-two-stage split. Two scoped checks indicate that the instability is not
confined to this basis. Locally rescoring a released public navhard two-stage
trajectory output matches its reported combined score under the same-source
control stack (41.74 versus 41.7) but returns 0.0 under the affected condition.
On 512 synthetic stage-2 frames, the affected condition diverges on 99.8--100\%
of rollouts versus 0\% under the control stack. Neither check evaluates the
final stage-1/stage-2 aggregate ranking across the audit agents, and synthetic stage-2 frames carry
no logged human trajectory, so reference-side forgiveness is not measurable
there.

Our two-sided finite-difference diagnostic replaces BatchLQR re-simulation for
both proposal and human reference with kinematics derived directly from poses.
It is not a replacement metric. Its symmetric intervention tests whether an
inversion depends on the shared refit rather than a one-sided advantage. Solver
controls hold the dependency condition, traffic, and forgiveness fixed while replacing only
the velocity fit solver with a direct solve or Hermitian pseudoinverse. The
fallback retains stock agent-side BatchLQR except when a rollout leaves its
proposal by more than $100$\,m or makes a step over $50$\,m, when it uses
finite-difference states. The NumPy-2.x control stack imports and runs unchanged scorer source. Thus
ordering changes isolate a dependency-sensitive numerical difference rather than
a scorer source revision.

The narrower fallback retains the raw human reference and stock agent-side
BatchLQR until a pre-specified gross divergence threshold is crossed. Agreement
supports divergence rather than wholesale BatchLQR removal as the explanation.

\subsection{Formal failure condition}
Let $x_a$ and $x_r$ be agent proposal and reference, $T(\cdot,s)$ their shared rollout and refit map in scene $s$, and $m_j(T(x),s)\in[0,1]$ the post-transform
score of channel $j$. The reference failure mask is
\[
F_j(x_r,s)=\mathbf{1}[m_j(T(x_r,s),s)=0],
\]
and stock filter-on scoring is
\[
\tilde m_j(x_a,x_r,s)=
\begin{cases}
1,&F_j(x_r,s)=1,\\
m_j(T(x_a,s),s),&\text{otherwise.}
\end{cases}
\]
For ranking-constraining channels $\mathcal D$, joint failure of $T$ with $F_j=1$
for most $j\in\mathcal D$ removes those structural constraints from the aggregate
and leaves residual terms to rank agents. The observable signature is therefore
a blind win under stock filter-on, fail-to-pass overwrite on structural channels,
and blind demotion when shared failure is removed or guarded. The premise fails
if reference scoring is independent or masks are localized rather than
overwriting agent channels.

This condition does not require every score channel to fail. It predicts loss of
the structural channels in $\mathcal D$ that would otherwise constrain unsafe or
non-compliant trajectories. The audit consequently reports both aggregate rankings and overwrite masks and rates because an aggregate reversal without the mask evidence would not establish the proposed mechanism.

\subsection{Agents and evaluation discipline}
\ignoreall is route-blind. It uses current ego speed and an empty actor set in a
free-road IDM rollout ($v_{\mathrm{des}}=15$, $a=1.5$, $b=3.0$), with a 4\,s
horizon sampled every 0.5\,s, no privileged path, and no logged future. The
second probe is route-aware yet actor-blind. It uses current ego state, ordered mission
roadblock IDs, and static-map lane and lane-connector baselines, with no
actor boxes, tracks, trajectories, collision predictions, or logged future ego
trajectory. All forbidden-input and second-process determinism checks pass on
12,146 frozen navtest tokens. A separate route boundary audit measures
route adherence and moving-start progress. It tests neither collision avoidance
nor safety.

Both probes are deterministic and use a single parameterization across all tokens and score bases, with no retuning for individual scenes or diagnostics. An independent boundary
audit reconstructs the route-aware probe from its declared inputs, obtains
byte-identical output in a second process and verifies all forbidden-input
checks. These checks establish an actor-free route-following diagnostic, not an
autonomous driving baseline.

We compare these probes with \humanreplay, \pdmc, \ddrive, and the released
LEAD planner's LTFV6 configuration and checkpoint, denoted
\ltfv~\cite{nguyen2026lead}. The
scorer always evaluates collision, time-to-collision, progress, and compliance
against true simulator state. We use reactive IDM traffic and non-reactive log
replay to test whether traffic yielding explains a blind win. All paired
confidence intervals use a log-clustered bootstrap. We
resample the 136 navtest logs with replacement and retain paired agent scores
within each sampled log (4,000 draws, seed 0). The solver matrix uses all 450 original scene stage-1 tokens from the official
interaction-focused navhard-two-stage split, with no further selection. The fresh-install experiment uses a deterministic 32-token failure reproduction
list whose original parent pool rule is unavailable. It is therefore treated only as
mechanism evidence, not as a prevalence sample. For reproducibility, code and frozen artifacts are available at
\url{https://github.com/WZiang/navsim-score-basis-audit}. The repository contains token lists,
environment specifications, exact-input arrays with SHA-256 hashes, a standalone solver
reproducer, and per-token outputs. The package versions, numerical backends, and thresholds
required to reproduce every reported condition are specified in this paper.
Margins and intervals use
unrounded per-token scores, whereas tabulated aggregates are rounded to one
decimal. The released checkpoint rescoring aligns all
rows on a separate 12,143-token intersection. These scopes are kept distinct throughout. 

\section{A NAVSIM Score Basis Audit}
\label{sec:results}

\subsection{The stock scorer rewards actor-blind probes}
\label{sec:results-rank}

\begin{table}[t]
\centering
\small
\setlength{\tabcolsep}{4.0pt}
\caption{\textbf{Full-navtest anomaly under the stock filter-on scorer.}
In the affected documented-stack condition, two actor-blind probes outrank actor-aware
comparators. The route-aware probe is evaluated on the frozen 12,146-token
universe; unavailable subset scores are omitted. Frozen bins are turn+moving
($n{=}3{,}204$), low-speed ($<2$\,m/s; $n{=}2{,}539$), and close-lead
(gap $<12$\,m, closing $>0.5$\,m/s; $n{=}574$). NC denotes no at-fault
collision, and TTC denotes time-to-collision. NC fail and TTC fail are percentages; all
other reported values are \epdms $\times100$. All rows use 12,146 tokens
except DiffusionDrive, which uses its 12,143-token decoded coverage.}
\label{tab:full-navtest}
\begin{tabular}{lcccccc}
\toprule
Agent & Full & \shortstack[c]{Turn+\\moving} & \shortstack[c]{Low-\\speed} & \shortstack[c]{Close-\\lead} & \shortstack[c]{NC\\fail} & \shortstack[c]{TTC\\fail} \\
\midrule
\ignoreall & \textbf{79.6} & \textbf{82.8} & \textbf{77.2} & \textbf{79.7} & \textbf{4.9} & \textbf{5.0} \\
\shortstack[l]{Route-aware\\actor-blind} & 79.2 & -- & -- & -- & -- & -- \\
Human replay & 74.0 & 79.2 & 70.4 & 68.8 & 10.7 & 11.2 \\
\ddrive & 73.7 & -- & -- & -- & -- & -- \\
\pdmc & 67.1 & 71.3 & 62.1 & 65.2 & 16.7 & 18.0 \\
\bottomrule
\end{tabular}
\end{table}

Under stock filter-on scoring, \ignoreall scores $79.6$ \epdms on full navtest,
above \humanreplay ($74.0$), \pdmc ($67.1$), and \ddrive ($73.7$). It also
exceeds human replay by $5.9$ points under non-reactive replay, so traffic
yielding does not explain the anomaly. The scorer remains responsive to the true scene actors, as evasive zero-velocity trajectories recover collision success.
Thus, this is a ranking anomaly under the stock score basis, not evidence that
blind driving is competitive.

The route-aware actor-blind probe removes the straight-path confound. It scores
$79.2$ over the complete 12,146-token navtest split, exceeding human replay by
$5.3$ points (95\% CI $[4.4, 6.2]$) and \pdmc by $12.1$ points
($[10.5, 13.7]$). Its $-0.4$-point margin against \ignoreall
($[-0.9, 0.2]$) is inconclusive, so we claim no stock score difference between the
two blind probes. A second stock run exactly reproduces all per-token score outputs and the
aggregate; only runtime metadata differs.

Inspected collision risk cases confirm that the score retains the true scene.
The anomaly is therefore not a hidden perception or simulator visibility
failure. The independent boundary audit passes all forbidden-input and second-process
determinism checks. Overall, 99.0\% of tokens remain within 4\,m of the intended
route, and 99.97\% of moving starts make positive progress. All 118 route tail cases
arise from terminal roadblocks with no outgoing edge followed by endpoint
extrapolation. Excluding them leaves the route-aware--human margin essentially
unchanged at $+5.3$ points (95\% CI $[4.4, 6.2]$). The full navtest estimate remains
primary.
The probe therefore rules out the simpler explanation that a fixed straight
path accidentally receives compliance credit, while making no claim about
collision avoidance or safety.

\subsection{Shared rollout failure rewrites the score basis}
\label{sec:results-mechanism}

NAVSIM zeroes waypoint velocities before re-simulation and refits velocity and
curvature through BatchLQR. The stock pseudoinverse can return an unphysical
profile whose rollout travels kilometers off-road. The coarse logged reference
shares this path. Reference-conditioned forgiveness then overwrites the
corresponding agent failures to passes, removing structural constraints that
would otherwise penalize invalid trajectories.

This instability is not forced by the trajectory poses. The current ego speed
is known and an accurate high-precision solution exists for the same regularized normal system,
yet the dependency-sensitive default pseudoinverse can select an implausible
profile.

\begin{table}[t]
\centering
\footnotesize
\setlength{\tabcolsep}{3.3pt}
\caption{\textbf{Score-basis rewrite under shared rollout failure.}
Scores are \epdms $\times100$ and rates are percentages. The diagnostic score
uses two-sided finite differences; core overwrite averages DAC, DDC, and LK channels.
The DiffusionDrive row uses its 12,143-token decoded coverage; all other rows use
12,146 tokens.}
\label{tab:score-basis-disclosure}
\begin{tabular}{@{}lccccc@{}}
\toprule
Agent & \shortstack[c]{stock\\score} & \shortstack[c]{diagnostic\\score} & \shortstack[c]{core overwrite\\rate (\%)} & \shortstack[c]{fallback\\rate (\%)} & \shortstack[c]{path dev. km\\med. / p90} \\
\midrule
\ignoreall & 79.6 & 13.3 & 100.0 & 100.0 & 68.4 / 195.5 \\
Human replay & 74.0 & 88.9 & 99.9 & 99.4 & 7.2 / 22.8 \\
\pdmc & 67.1 & 86.2 & 99.7 & 99.5 & 4.2 / 12.5 \\
\ddrive & 73.7 & 85.3 & 99.9 & 99.6 & 7.1 / 21.9 \\
\bottomrule
\end{tabular}
\end{table}

Only stock filter-on combines shared failure with forgiveness and yields a
positive blind-over-human margin. Removing or guarding the failure restores
blind-last ordering.

The same reference failure mask rewrites core compliance channels for almost
every audited token, even when agents follow different proposals. It removes
structural checks that would otherwise constrain the aggregate, allowing
residual terms to invert the blind policy ordering. The learned-agent row
exhibits the same instability and overwrite pattern, showing that the mechanism
is not confined to constructed probes.
For \ignoreall specifically, raw failures become passes on 99.98\% of DAC,
99.98\% of DDC, and 99.92\% of lane keeping evaluations, compared with only
2.63\% for TTC and 2.39\% for no at-fault collision. A fail-to-pass overwrite
requires failure on both the agent and reference sides. DAC, DDC, and lane keeping
compare the refitted rollout directly with road and lane geometry, so
kilometer-scale divergence broadly fails these structural channels on both
sides. NC and TTC instead require a conflict with a particular surrounding
actor. A divergent rollout can move away from actors rather than intersect
them. The rewrite is therefore concentrated in structural channels, not a
uniform inflation of every metric.

\begin{table}[t]
\centering
\footnotesize
\setlength{\tabcolsep}{2.5pt}
\caption{\textbf{Ingredient-removal causal check.}
The blind advantage requires both shared rollout failure and
reference-conditioned forgiveness. Removing or guarding the failure restores
blind-last ordering. Scores and margins are \epdms $\times100$. Human denotes
human replay, PDM denotes \pdmc, and IA denotes \ignoreall. Near-zero
filter-off values are shown to two decimal places.}
\label{tab:ingredient-causal-check}
\begin{tabular}{@{}p{0.21\textwidth}ccccccl@{}}
\toprule
Basis & \shortstack[c]{shared\\failure} & forgiveness & Human & PDM & \IA & \shortstack[c]{IA$-$Human} & result \\
\midrule
Stock filter-on & yes & on & 74.0 & 67.1 & 79.6 & +5.6 & collapse \\
Stock filter-off & yes & off & 0.00 & 0.09 & 0.00 & +0.00 & exposed \\
Two-sided & removed & on & 88.9 & 86.2 & 13.3 & -75.6 & blind-last \\
Fallback & guarded & on & 88.3 & 86.0 & 13.3 & -75.1 & blind-last \\
\bottomrule
\end{tabular}
\end{table}

The ingredient ablation separates failure exposure from behavioral ranking.
With forgiveness disabled, widespread rollout failures appear as near-zero
scores instead of passes. Although not useful as a leaderboard, this verifies
that the blind advantage requires the reference-conditioned rewrite. Keeping
forgiveness while removing or guarding the shared failure yields the opposite
ordering, showing that neither ingredient alone explains the stock result.

\begin{table}[t]
\centering
\footnotesize
\setlength{\tabcolsep}{1.2pt}
\caption{\textbf{Dependency-sensitive reversal on the 450-token control pool.}
Changing only the helper solver, or using the unchanged scorer in the same-source
control stack, removes divergence and restores blind-last ordering.
The frozen pool is the complete original scene stage-1 token set of the
official navhard-two-stage split (all 450 tokens valid); the split is
benchmark-curated toward interactive scenes and we apply no further selection.
Affected denotes the audited documented-stack condition used for the 450-token
solver matrix; Control denotes the same-source dependency stack control; Herm.\
pinv. is Hermitian pseudoinverse;
and IA denotes \ignoreall. Human-ref.\ fail is the fraction of tokens on which
human replay itself fails DAC, DDC, or lane keeping under filter-off scoring
with the same solver.}
\label{tab:stack-solver-matrix}
\begin{tabular}{@{}>{\raggedright\arraybackslash}p{0.20\textwidth}cccccc@{}}
\toprule
& \multicolumn{4}{c}{Scores (\epdms $\times100$)} & \multicolumn{2}{c}{Divergence} \\
\cmidrule(lr){2-5}\cmidrule(lr){6-7}
Setting & \IA & Human & \ltfv & \pdmc & \shortstack[c]{div.\\rate} & \shortstack[c]{human-ref\\fail} \\
\midrule
Affected + stock pinv. & 81.8 & 78.8 & 75.7 & 75.5 & 99.8\% & 100.0\% \\
Affected + direct solve & 43.9 & 62.8 & 53.1 & 60.4 & 0\% & 12.9\% \\
Affected + Herm.\ pinv. & 45.7 & 66.1 & 50.5 & 59.2 & 0\% & 9.1\% \\
Control stack + stock pinv. & 12.5 & 98.8 & 81.2 & 87.6 & 0\% & 0.0\% \\
\bottomrule
\end{tabular}
\end{table}

\begin{table}[t]
\centering
\footnotesize
\setlength{\tabcolsep}{3.4pt}
\caption{\textbf{Fresh-install mechanism reproduction on a fixed 32-token diagnostic set.}
This fixed diagnostic set is used only for mechanism reproduction because its
original sampling rule is unavailable; it is not treated as a prevalence sample.
Affected denotes the fresh installation of the documented stack; Control denotes its
same-source dependency stack control. On and off indicate whether reference-conditioned forgiveness is enabled. IA denotes the route-blind \ignoreall
probe. Human-ref. fail denotes failure in any of DAC, DDC, or LK on human replay,
and IA overwrite denotes any fail-to-pass overwrite in these core channels for IA.
Scores are \epdms on the $[0,1]$
scale.}
\label{tab:clean-chain-smoke}
\begin{tabular}{lccccc}
\toprule
Setting & \shortstack[c]{IA rollout\\div.} & \shortstack[c]{Human-ref.\\fail} & \shortstack[c]{IA\\overwrite} & IA score & Human score \\
\midrule
Affected, on  & 1.000 & 1.000 & 1.000 & 0.857 & 0.771 \\
Affected, off & 1.000 & 1.000 & 0.000 & 0.000 & 0.000 \\
Control, on & 0.000 & 0.156 & 0.125 & 0.125 & 0.990 \\
\bottomrule
\end{tabular}
\end{table}

On the frozen 450-token pool (the complete original-scene stage-1 set of the
official navhard-two-stage split), direct solve, Hermitian
pseudoinverse, and the same-source control stack all remove replay
divergence and demote \ignoreall. A fresh-install mechanism reproduction then
gives divergent rollouts on all 32 tokens, versus none for the same-source control stack.
The scorer and helper source hashes are identical. \Cref{tab:clean-chain-smoke}
shows, on the same deterministic diagnostic set, overwrite under Affected with
forgiveness enabled, failure exposure under Affected with forgiveness disabled,
and a ranking reversal under Control with forgiveness enabled. Its fixed
list is scoped mechanism evidence, not a prevalence estimate. Both start from
a median planned endpoint of $19.58$\,m. The affected documented-stack condition instead reaches a
median simulated endpoint of $5.13$\,km and path length of $32.11$\,km, whereas
the control remains at $19.72$\,m and $20.41$\,m, respectively.

An exact-input diagnostic loads one frozen $40\!\times\!40$ normal system in
both environments. Under the documented condition, the default pseudoinverse
violates the pre-specified fitted-state bounds ($v_0=-491.34$\,m/s,
$\max|a|=3.06\!\times\!10^5$\,m/s$^2$) and has a large residual, whereas the
Hermitian and NumPy-2 controls agree with the high-precision solution
($v_0\approx12.53$\,m/s, $\max|a|=1.04$\,m/s$^2$). Direct solve removes the rollout divergence, although it is not an accuracy anchor. This separates identical
input from dependency-sensitive solver behavior without attributing a
library-internal implementation cause. 
The solver matrix holds traffic and forgiveness fixed. Stabilizing the solver
reduces the human reference failure rate from $100.0\%$ to $12.9\%$ (direct
solve) and $9.1\%$ (Hermitian pseudoinverse), eliminates replay divergence,
and restores blind-last ordering on this control pool. Agreement across direct
solve, Hermitian pseudoinverse, and the dependency control localizes the
trigger. Although absolute scores vary with reconstructed kinematics, all three
remove replay divergence and recover blind-versus-reference ordering.

\subsection{A two-sided diagnostic restores blind-last ordering}
\label{sec:results-twosided}

\begin{table}[t]
\centering
\small
\setlength{\tabcolsep}{3.5pt}
\caption{\textbf{Two-sided control restores blind-last ordering.}
Scores are \epdms $\times100$. Applying the diagnostic to proposal and
reference places human replay and \pdmc above the route-aware actor-blind probe,
which in turn ranks above \ignoreall. Route-aware non-reactive values are
unavailable; DiffusionDrive uses matched decoded tokens. react. and non-react.
denote reactive and non-reactive traffic, respectively.}
\label{tab:twosided}
\begin{tabular}{@{}lccccc@{}}
\toprule
& \multicolumn{2}{c}{\shortstack[c]{Stock\\filter-on}} & \multicolumn{2}{c}{\shortstack[c]{Two-sided\\finite-diff.}} & Fallback \\
\cmidrule(lr){2-3}\cmidrule(lr){4-5}\cmidrule(lr){6-6}
Agent & react. & non-react. & react. & non-react. & react. \\
\midrule
Human replay & 74.0 & 73.3 & \textbf{88.9} & \textbf{98.0} & \textbf{88.3} \\
PDM-Closed & 67.1 & 66.5 & 86.2 & 86.1 & 86.0 \\
\shortstack[l]{Route-aware\\actor-blind} & 79.2 & -- & 45.1 & -- & 45.1 \\
\ignoreall & \textbf{79.6} & \textbf{79.2} & 13.3 & 15.0 & 13.3 \\
\midrule
DiffusionDrive & 73.7 & 73.4 & 85.3 & 92.3 & 84.9 \\
\bottomrule
\end{tabular}
\end{table}

Applying finite-difference kinematics symmetrically places human replay and
\pdmc above the route-aware actor-blind probe, which in turn ranks above
\ignoreall. The route-aware probe
scores $45.1$, placing it $43.8$ points below human replay and $31.8$ points
above \ignoreall. Fallback agrees, but serves only as corroborating evidence
because it activates on every token for this probe. Applying the same intervention to proposal and reference tests the
shared transformation without giving either side an asymmetric advantage.

For the original route-blind probe, the paired margin against human replay
changes from $+5.6$ [$4.8,6.6$] under stock filter-on to $-75.6$
[$-78.0,-73.2$] two-sided. The fallback yields $-75.1$ [$-77.6,-72.6$]. Similar
reversals hold against \pdmc, and on the matched decoded token subset
\ddrive also moves well above \ignoreall. These diagnostic interventions are
not replacement leaderboards, but they show that stabilization retains
route-following information while restoring the actor-blind distinction. Across four
fallback thresholds, from strict $(20\,\mathrm m,10\,\mathrm m)$ to loose
$(200\,\mathrm m,100\,\mathrm m)$ path and step limits, \ignoreall remains last.
The conclusion is not tied to the selected $100\,\mathrm{m}$ path and $50\,\mathrm{m}$ step guard.

\subsection{Diagnostics reverse the blind probe ordering for released checkpoints}
\label{sec:results-realagent}

\begin{table}[t]
\centering
\small
\setlength{\tabcolsep}{3.5pt}
\caption{\textbf{Released-checkpoint rescoring on the matched navtest intersection.}
Stock scoring ranks \ignoreall first. Both diagnostics rank every comparator
above \ignoreall. Values are \epdms $\times100$; ranks are in parentheses.
All rows use the same matched intersection ($n{=}12{,}143$).}
\label{tab:consequence-audit}
\begin{tabular}{lccc}
\toprule
Agent & stock filter-on & two-sided & fallback \\
\midrule
\ignoreall    & 79.6 (1) & 13.3 (5) & 13.3 (5) \\
\humanreplay  & 74.0 (2) & 88.9 (1) & 88.3 (1) \\
\ddrive & 73.7 (3) & 85.3 (3) & 84.9 (3) \\
\pdmc         & 67.1 (4) & 86.2 (2) & 86.0 (2) \\
\ltfv         & 0.0 (5)  & 84.4 (4) & 83.4 (4) \\
\bottomrule
\end{tabular}
\end{table}

All rows in \Cref{tab:consequence-audit} use the same matched 12,143-token
intersection. The \ltfv row uses the released \ltfv configuration and
checkpoint from LEAD~\cite{nguyen2026lead}. We convert its CARLA-frame waypoints and
headings to ISO~8855. Stock scoring ranks \ignoreall first and \ltfv at zero,
whereas both diagnostics rank every comparator above \ignoreall. This reversal
diagnoses the score-basis inversion. The diagnostic scores are not intended as a
replacement leaderboard.

The checkpoint covers 12,143 of 12,146 tokens. Three truncated-image decode
failures are omitted. Coordinate-range checks pass, and the retained decoded
trajectories recover strong diagnostic scores. This does not validate every
exporter configuration, so the \ltfv row remains supporting evidence. On the
matched 574-token close-lead pool, \ignoreall is worst by open-loop ADE
($7.63$\,m versus $0.54$\,m for \ltfv) while winning stock score. After overwrite,
no convex reweighting of the component scores gives \pdmc more than a $0.17$-point margin
over \ignoreall. Thus, neither logged-future imitation nor poor aggregate
weighting explains the inversion. Discrimination is already lost from the score basis.

\section{Implications for Defensive Driving Evaluation}
\label{sec:mechanism}

The dependency-sensitive velocity refit instability is the direct trigger in
the audited configuration. Reference-conditioned forgiveness does not cause
that instability; it propagates shared reference failures into agent credit and
enables the observed ranking inversion. Keeping forgiveness enabled while
stabilizing the solver removes divergence and restores blind-last ordering on
the 450-token control pool. Blind-last recovery is only a necessary property, since a score that cannot distinguish an actor-aware policy from one that ignores surrounding actors cannot support stronger behavioral claims.

Collapse requires a common transformation applied to proposal and reference,
forgiveness conditioned on reference failure, and broad shared failure in
behavior-relevant channels. Stable rollouts remove the observed collapse, but
reference-conditioned forgiveness still requires stability checks because it
has no independent mechanism to distinguish shared numerical failure from a
legitimate reference exception. The symmetric two-sided intervention tests the
common transformation without favoring either side, while fallback retains the
stock path until gross divergence.

Forgiveness still addresses real false positives, as strict no-forgiveness scoring can drive \humanreplay and \pdmc to zero on interaction-heavy candidates. We therefore recommend reporting dependency versions and numerical backend, reference failure masks and fail-to-pass overwrite rates, route-blind and route-aware actor-blind probes, and fitted-state and rollout stability before interpreting rankings.
 These checks do not certify defensive driving. A blind win instead exposes failure of
a necessary condition for actor-sensitive interpretation.

\section{Limitations}
\label{sec:limitations}

\paragraph{Empirical scope.}
We audit pseudo-simulation, not closed-loop driving. The complete signature is
established for NAVSIM original scene single-stage scoring in the affected
documented-stack condition on our x86-64 backend with OpenBLAS. Blind probe results cover 12,146 tokens from the full navtest set, whereas released-checkpoint rescoring is performed on the matched intersection of 12,143 tokens. The 450-token control pool, 32-token diagnostic set, and exact-input case serve as scoped controls. They support the audited
configuration but neither identify a library-internal cause nor imply failure
across branches, platforms, or backends.
Because published evaluations rarely disclose SciPy and BLAS/LAPACK
configurations, we cannot determine how often this condition
occurred in reported results.

\paragraph{Protocol scope.}
We rescore one released two-stage output as a scoped stability check, but do
not evaluate the final stage-1/stage-2 aggregate across the audit agents. We
therefore draw no conclusion about the prevalence or magnitude of this failure
in official leaderboard results. The checks
in \cref{sec:audit} support the relevance of rollout stability beyond the
primary score basis, but do not establish aggregate ranking collapse.

\paragraph{Interpretation.}
Actor sensitivity is necessary but not sufficient. Route boundary checks show
input hygiene and route competence, not safety, and blind-last recovery restores
only the distinction under audit. The controls are diagnostics, not production
or deployment metrics. Future work should apply the same blind baseline
and stability audit to more submissions and re-simulation benchmarks.

\section{Conclusion}
\label{sec:conclusion}

We identify a dependency-sensitive numerical failure in the shared NAVSIM
rollout and refit path. Reference-conditioned forgiveness does not cause this
instability, but propagates the resulting reference failures into broad agent
credit, enabling the observed ranking inversion. Solver, stack, and score basis
controls remove the collapse and restore blind-last ordering.
These findings apply only to the audited configuration. They motivate dependency disclosure, overwrite reporting, blind probes, and rollout stability checks.



\bibliographystyle{splncs04}
\bibliography{main}

\end{document}